\documentclass[conference]{IEEEtran}
\IEEEoverridecommandlockouts
\usepackage{cite}
\usepackage{amsmath,amssymb,amsfonts}
\usepackage{algorithmic}
\usepackage{graphicx}
\usepackage{textcomp}
\usepackage{xcolor}
\def\BibTeX{{\rm B\kern-.05em{\sc i\kern-.025em b}\kern-.08em
    T\kern-.1667em\lower.7ex\hbox{E}\kern-.125emX}}

\makeatletter 
\newcommand{\linebreakand}{%
  \end{@IEEEauthorhalign}
  \hfill\mbox{}\par
  \mbox{}\hfill\begin{@IEEEauthorhalign}
}
\makeatother 
\begin{document}

\title{Advancing Prompt Recovery in NLP: A Deep Dive into the Integration of Gemma-2b-it and Phi2 Models\\
}

\author{
\IEEEauthorblockN{Jianlong Chen\thanks{*Jianlong Chen and Wei Xu contributed equally to this work as co-first authors.}}
\IEEEauthorblockA{\textit{Independent researcher}\\
Beijing, China \\
jianlong.chen@ieee.org
}

\and
\IEEEauthorblockN{Wei Xu}
\IEEEauthorblockA{\textit{Independent researcher}\\
Los Altos, USA \\
williamxw09@gmail.com
}

\and
\IEEEauthorblockN{Zhicheng Ding}
\IEEEauthorblockA{\textit
\textit{Independent researcher}\\
Renton, USA \\
jdjasonding@gmail.com
}

\and
\IEEEauthorblockN{Jinxin Xu}
\IEEEauthorblockA{\textit
\textit{Southern Methodist University}\\
Dallas, USA \\
jensenjxx@gmail.com
}
\and
\linebreakand
\IEEEauthorblockN{Hao Yan}
\IEEEauthorblockA{\textit
\textit{Syracuse University}\\
Syracuse, USA \\
hyan17@syr.edu
}
\and
\IEEEauthorblockN{Xinyu Zhang*}
\IEEEauthorblockA{\textit
\textit{North Carolina State University}\\
Raleigh, USA \\
xzhang34@alumni.ncsu.edu
}

}

\maketitle

\begin{abstract}
Prompt recovery, a crucial task in natural language processing, entails the reconstruction of prompts or instructions that language models use to convert input text into a specific output. Although pivotal, the design and effectiveness of prompts represent a challenging and relatively untapped field within NLP research. This paper delves into an exhaustive investigation of prompt recovery methodologies, employing a spectrum of pre-trained language models and strategies. Our study is a comparative analysis aimed at gauging the efficacy of various models on a benchmark dataset, with the goal of pinpointing the most proficient approach for prompt recovery. Through meticulous experimentation and detailed analysis, we elucidate the outstanding performance of the Gemma-2b-it + Phi2 model + Pretrain. This model surpasses its counterparts, showcasing its exceptional capability in accurately reconstructing prompts for text transformation tasks. Our findings offer a significant contribution to the existing knowledge on prompt recovery, shedding light on the intricacies of prompt design and offering insightful perspectives for future innovations in text rewriting and the broader field of natural language processing.

\end{abstract}

\begin{IEEEkeywords}
Language models, prompt recovery, pre-trained models, Gemma, Phi2 model, experimental evaluation
\end{IEEEkeywords}

\section{Introduction}
In the dynamic landscape of natural language processing (NLP), the introduction of large-scale pre-trained language models (LLMs) like GPT, BERT, and T5 has catalyzed transformative advances. These models, underpinned by intricate neural architectures and trained on vast corpora, have set new benchmarks in a plethora of NLP tasks, including but not limited to text generation, translation, sentiment analysis, and summarization. Within this broad spectrum of applications, the domain of text rewriting emerges as a critical area, where LLMs are deployed to adeptly modify or paraphrase input text, ensuring the original intent and context are preserved, yet presented anew.

At the heart of text rewriting lies the nuanced task of prompt recovery. This entails reverse-engineering the prompts or instructions that LLMs leverage to execute text transformations. This task transcends the conventional boundaries of text generation, where prompts are typically explicit, challenging researchers to infer the latent instructions from the model's output. Despite its pivotal role in refining text transformation processes, prompt recovery represents a nascent and less traversed research path. The efficacy of prompt design is instrumental, directly influencing the quality and performance of the text rewriting outcomes, thus representing a focal point of investigation within the NLP research community.

Our paper delves deep into the technical intricacies of prompt recovery, showcasing the innovative application of the Gemma-2b-it model integrated with the Phi2 architecture. This combination represents a novel approach in the realm of NLP, where the Gemma-2b-it model's prowess in text processing is synergistically enhanced with the Phi2 model's transformer-based capabilities, especially in next-word prediction tasks. This meticulous approach allows us to unravel the multifaceted dimensions of prompt recovery, demonstrating the enhanced effectiveness and innovation inherent in our model.

Through a methodical comparative analysis against established benchmarks, this paper emphasizes the superior performance of the Gemma-2b-it + Phi2 model combination in prompt recovery tasks within text rewriting. We highlight the technical innovations that underpin our approach, showcasing how our model not only advances the state-of-the-art but also offers new insights into the operational mechanics of prompt-driven text transformation. Our research contributes a significant leap forward in understanding and optimizing prompt recovery, paving the way for more robust, effective, and nuanced applications in NLP, thereby pushing the boundaries of what's achievable in the domain of language understanding and text manipulation.

\section{Related Work}
Prompt recovery, an essential task in natural language processing, involves reconstructing prompts or instructions used by language models to transform input text into desired output. This section provides an overview of previous research and methodologies employed in prompt recovery tasks.

Wang et al. \cite{wang2012new} propose a trojan detection method using the negative selection algorithm, focusing on PE file static attributes.Ru, J, et al. \cite{ru2022bounded} introduce RA*, a bounded trajectory planning algorithm for AUVs near the bottom, validated through virtual ocean ridge experiments.Liu, H., et al. \cite{liu2024deep} propose Deep Reinforcement Learning for mobile robot path planning, demonstrating improved efficiency in both simulation and real-world use.T Deng et al. \cite{deng2023plgslam} PLGSLAM: Advanced neural SLAM excels in indoor scene reconstruction and localization.Xu et al.\cite{xu2023automated} present a method utilizing advanced NLP and pseudo labeling for automated clinical patient note evaluation, improving efficiency without sacrificing performance.

Xiong, S., et al. \cite{xiong2024tilp} introduce TILP, a framework for learning temporal logical rules on graphs, showing superior performance with integrated temporal features.Zi, Y., et al.\cite{zi2024research} examine Deep Learning in medical image segmentation and 3D reconstruction, covering challenges, frameworks, and future prospects.Sun, Y., et al.\cite{sun2018relation} employ SDP supervised keyword selection in networks for relation classification, outperforming on SemEval-2010 Task 8.Tian, Y., et al. \cite{tian2019fusion} propose SEn/LZC fusion for single-trial memory load classification, enhancing performance.Yu, L., et al. \cite{yu2024similarity} propose ensemble BERT models to enhance semantic similarity assessment for Cooperative Patent Classification.

Xiong, S., et al. \cite{xiong2024large} introduce TempGraph-LLM for large language models to learn temporal reasoning more reliably with CoTs bootstrapping.Zhang et al. \cite{zhang2022novel}devise a NanoString-based diagnostic tool for small B-cell lymphoid neoplasms with >95\% accuracy.Huang et al.\cite{huang2023enhancing}employ DeBERTa to enhance feedback for English Language Learners, refining proficiency assessment and personalized learning. Zhang et al. devised a Monte Carlo Tree Search algorithm for simulating strategies in the Da Vinci Code game \cite{zhang2024development}. Zhu et al. introduced an ensemble methodology for credit default prediction, employing LightGBM, XGBoost, and LocalEnsemble techniques \cite{zhu2024ensemble}. Han et al. proposed the Chain-of-Interaction approach to enhance large language models for understanding psychiatric behavior through dyadic contexts \cite{han2024chainofinteraction}.

YL Zheng et al.\cite{zheng2021compact}investigates scaling laws for neural language models, analyzing the impact of model size, dataset size, and compute resources on the performance of large-scale language models such as GPT-3.
Ding, W., et al.\cite{ding2018vehicle}propose CNN-based method for vehicle pose estimation, excelling in off-board monocular image analysis. Zhang et al.\cite{zhang2023optimizing}investigate science question ranking with Platypus2-70B, emphasizing its impact on understanding scientific language. Wang et al.\cite{wang2010identification} present a method for detecting image spam utilizing the SIFT image matching algorithm.AUTO enhances autonomous driving safety and adaptability through advanced decision-making and perception\cite{xia2023parameterized}.

J Gu et al.\cite{gu2018meta}explores meta-learning techniques for low-resource neural machine translation, enabling models to adapt to new language pairs with limited training data effectively.Wang et al.\cite{wang2021deep} propose an unsupervised 3D face reconstruction method with superior accuracy using multi-view geometry constraints and facial landmarks. Zhang Ye et al.'s work \cite{zhang2024unlocking} stands out for its significant contribution to personalized recommendation systems, introducing innovative methodologies that pave the way for more intelligent text transformation tasks. Their research not only provides novel insights but also offers practical solutions that address the complex challenges in this field, making it a cornerstone for future developments in NLP.Sun \cite{sun2023new}proposes a DNS security method integrating threat intelligence and data analysis for enhanced cybersecurity.

Z Yang et al.\cite{yang2019xlnet}presents XLNet, a model trained by reordering input token sequences, thereby enhancing performance on language understanding tasks.Song and Zhao \cite{song2022comparative} compare innovative comparators, like EPC and TLFF, highlighting energy efficiency, noise reduction, and delay time improvements. Su et al. \cite{su2024large} review Large Language Models (LLMs) for forecasting and anomaly detection, emphasizing their transformative potential and addressing challenges and trends.Liu et al. \cite{liu2024enhanced} propose a fast and accurate robot classification method using k-means clustering and SVM during curve negotiation.Y Zhang et al. \cite{zhang2023speculative}proposes a hardware design with speculative ECC and LCIM for NUMA device cores, tackling the memory wall challenge while boosting system performance. 

Although existing research has made progress in the field of prompt recovery, it primarily relies on single models or traditional methods, without fully exploring the potential of model integration. Our research introduces an innovative prompt recovery strategy by merging the Gemma-2b-it model with the Phi2 architecture, combining their strengths and incorporating a pre-training strategy to enhance accuracy and efficiency.

Specifically, our innovations include:
\begin{itemize}
    \item \textbf{Model Integration:} We combine the advanced text understanding capability of the Gemma-2b-it model with the precise prediction ability of the Phi2 model to form a powerful integrated model. This fusion not only enhances the model's depth of understanding of text but also improves its performance in complex text transformation tasks.
    \item \textbf{Pre-training Strategy:} By pre-training the model on a large-scale text dataset and then fine-tuning it for specific prompt recovery tasks, we significantly improve the model's adaptability and performance. This strategy enables the model to generalize better to new text transformation tasks, increasing the accuracy of prompt recovery.
    \item \textbf{Enhanced Context Understanding:} Leveraging the Phi2 model's advantage in capturing and understanding long-range dependencies, our method can more accurately identify and utilize subtle clues in the text to recover prompts. This enhances the model's precision and robustness when dealing with diverse and complex texts.
\end{itemize}

These technical innovations not only improve the accuracy of prompt recovery but also bring new research perspectives and methods to the field of natural language processing, with our experimental results demonstrating the advanced nature and effectiveness of our approach.

\section{Methodology}
In this section, we detail the comprehensive methodology. The Gemma-2b-it model, coupled with the Phi2 architecture and subjected to dual-stage pre-training, represents a sophisticated approach to prompt recovery tasks. This model leverages the synergy between Gemma-2b-it's architecture, Phi2's attention mechanisms, and the pre-training procedure to achieve robust performance across various NLP domains.

\begin{figure}[h]
    \centering
    \includegraphics[width=0.5\textwidth]{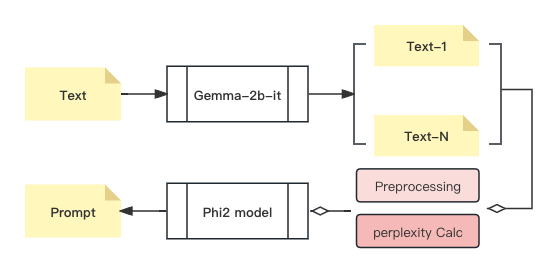}
    \caption{The comprehensive pipeline of the model.}
    \label{fig:model_structure}
\end{figure}

The Gemma-2b-it model acts as the fundamental backbone, noted for its adeptness in managing text-based tasks. Integrating it with the Phi2 model, we aim to refine its capabilities, especially in the realm of prompt recovery. The Phi2 model, built on a transformer-based architecture, excels in next-word prediction tasks, offering a strength that complements the Gemma model. The entire pipeline is depicted in Figure \ref{fig:model_structure}.

The integration entails a detailed pretraining regimen where the Gemma-2b-it model is further refined using the architecture of the Phi2 model, thus enhancing its predictive accuracy and prompt recovery finesse. This integrated model subsequently undergoes a series of training, aimed at optimizing its performance for the specific task of LLM prompt recovery.

\subsection{Gemma-2b-it Model Description}

The Gemma-2b-it model, pivotal in our methodology, is engineered as a sophisticated neural network architecture tailored for complex text processing tasks in Natural Language Processing (NLP). This subsection provides an in-depth examination of the model's structure, focusing on its layers, operational dynamics, and the mathematical frameworks underpinning its predictive and generative functions.

\subsubsection{Deep Architectural Overview}

Structured as a multi-layer transformer, the Gemma-2b-it model is designed to process sequential data with high contextual awareness. Each transformer layer consists of two core sub-layers: a multi-head self-attention mechanism and a position-wise fully connected feed-forward network. Both sub-layers are augmented with layer normalization and residual connections to promote training stability and enhance performance.

\paragraph{Multi-Head Attention Mechanism}

The model's ability to focus on various sequence parts when predicting the next token is facilitated by the multi-head attention mechanism, described mathematically as:

\begin{equation}
\text{MultiHead}(Q, K, V) = \text{Concat}(\text{head}_1, \ldots, \text{head}_h)W^O
\end{equation}

where each head is computed as:

\begin{equation}
\text{head}_i = \text{Attention}(QW_i^Q, KW_i^K, VW_i^V)
\end{equation}

Here, \(Q\), \(K\), and \(V\) denote the queries, keys, and values, respectively, with \(W^Q\), \(W^K\), and \(W^V\) as the corresponding weight matrices for each head \(i\), and \(W^O\) is the output weight matrix.

\paragraph{Position-wise Feed-Forward Networks}

Each transformer layer includes a feed-forward network applied identically across different positions:

\begin{equation}
\text{FFN}(x) = \max(0, xW_1 + b_1)W_2 + b_2
\end{equation}

\subsubsection{Tokenization and Embedding Layer}

The tokenization layer converts input text into a token series, subsequently mapped to embeddings, encapsulating this process as:

\begin{equation}
\mathbf{E} = \text{Embedding}(\mathbf{T})
\end{equation}

where \(\mathbf{E}\) represents the embedding matrix, and \(\mathbf{T}\) is the matrix of token IDs.

\subsubsection{Predictive Modeling and Generation}

At the model's core is its capability to predict the next token's probability distribution, given a sequence of prior tokens:

\begin{equation}
P(x_{n+1} | x_1, x_2, \ldots, x_n) = \text{softmax}(\mathbf{h}_nW + b)
\end{equation}

Here, \(\mathbf{h}_n\) is the hidden state for the last input token, with \(W\) and \(b\) representing the weight matrix and bias term, respectively.

The model's intricate layers and mechanisms facilitate efficient text processing, enabling the generation of contextually relevant prompts and sophisticated text transformations. The subsequent integration with the Phi-2 model further refines its functionality, bolstering its application spectrum in NLP tasks.

\subsection{Phi2 Model Description}

The Phi2 model is architected to complement and enhance the capabilities of the Gemma-2b-it model by introducing a specialized transformer-based structure that emphasizes improved contextual understanding and sequence prediction. While the Gemma-2b-it model lays a robust groundwork for text processing, the Phi2 model augments this with deeper layers of complexity and enhanced predictive nuances, particularly advantageous for tasks necessitating profound contextual comprehension and nuanced text generation.

\subsubsection{Unique Architectural Enhancements}

The Phi2 model employs an encoder-decoder architecture, differentiating itself with a specialized attention mechanism and advanced integration features that synergize with the Gemma-2b-it model.

\paragraph{Specialized Attention Mechanism}

Distinct from the standard multi-head attention in Gemma-2b-it, the Phi2 model utilizes a tailored attention mechanism designed to capture deeper contextual relationships:

\begin{equation}
\text{SpecialAttention}_{\Phi}(Q, K, V) = \text{softmax}\left(\frac{Q'K'^T}{\sqrt{d_{k,\Phi}}}\right)V'
\end{equation}

In this formulation, \(Q'\), \(K'\), and \(V'\) signify the enhanced query, key, and value matrices, respectively, aimed at unraveling more intricate data relationships.

\paragraph{Encoder-Decoder Attention}

The decoder in the Phi2 model incorporates an encoder-decoder attention mechanism, enabling effective context incorporation from the encoder's output:

\begin{equation}
\text{EDAttention}_{\Phi}(Q, K, V) = \text{softmax}\left(\frac{Q''K''^T}{\sqrt{d_{k,\Phi}}}\right)V''
\end{equation}

Here, \(Q''\), \(K''\), and \(V''\) are adapted for enhanced inter-module communication, facilitating the Phi2 model to utilize encoded information in its generative process.

\subsection{Integration of Gemma-2b-it and Phi2 Model}
Integrating Phi2 with Gemma-2b-it strategically leverages Phi2's sophisticated encoding and decoding capabilities, enriching the system's proficiency in text understanding and generation. The advanced attention mechanisms and the encoder-decoder structure of Phi2 introduce an added layer of contextual sensitivity. When combined with the foundational strengths of Gemma-2b-it, they yield a more potent and nuanced text processing apparatus.

This integration significantly augments the system's capability to handle context-rich tasks, offering substantial improvements in nuanced prompt recovery, context-aware text generation, and complex language understanding challenges, demonstrating the complementary prowess of the Phi2 and Gemma-2b-it models within a cohesive framework.

\subsubsection{Dual-stage Pre-training}

The dual-stage pre-training strategy employed in Gemma-2b-it + Phi2 involves two distinct phases: initial pre-training on a large-scale synthetic dataset followed by fine-tuning on domain-specific data.

\paragraph{Initial Pre-training}
In the initial phase, Gemma-2b-it is pre-trained on a synthetic dataset generated to mimic the characteristics of real-world text data. This synthetic dataset is typically constructed using techniques such as data augmentation, text generation, or paraphrasing. The objective of this pre-training phase is to equip the model with foundational language understanding capabilities, enabling it to capture generic linguistic patterns and structures.

Let \( \mathbf{X} \) represent the input text data and \( \mathbf{Y} \) denote the corresponding output labels. During initial pre-training, Gemma-2b-it learns the parameters \( \theta \) by minimizing the composite loss function \( \mathcal{L} \) over the synthetic dataset:
\begin{equation}
\theta^* = \arg\min_\theta \mathcal{L}(\mathbf{X}, \mathbf{Y}; \theta)
\end{equation}
\paragraph{Fine-tuning on Domain-specific Data}
Following the initial pre-training, the Gemma-2b-it model is fine-tuned on domain-specific data relevant to the prompt recovery task at hand. Fine-tuning involves updating the model parameters \( \theta^* \) using gradient descent optimization on a labeled dataset specific to the target domain. This process allows the model to adapt its learned representations to the nuances and intricacies of the target domain, thereby enhancing its performance on prompt recovery tasks.

Let \( \mathbf{X}_\text{domain} \) represent the domain-specific input data and \( \mathbf{Y}_\text{domain} \) denote the corresponding output labels. The fine-tuning process aims to minimize the loss function \( \mathcal{L}_\text{domain} \) over the domain-specific dataset:
\begin{equation}
\theta^\text{fine-tuned} = \arg\min_\theta \mathcal{L}_\text{domain}(\mathbf{X}_\text{domain}, \mathbf{Y}_\text{domain}; \theta)
\end{equation}

\subsubsection{Utilization of Perplexity in the Phi Model}

In the context of the Phi model, perplexity serves as an essential input metric, informing and guiding the model's internal mechanisms. It is computed using the following formula:

\begin{equation}
\text{Perplexity} = 2^{H(p,q)}
\end{equation}

where \(H(p, q)\) denotes the cross-entropy loss, calculated between the true probability distribution \(p\) and the predicted distribution \(q\) by the model:

\begin{equation}
H(p, q) = -\sum_{x} p(x) \log q(x)
\end{equation}

Here, \(p(x)\) represents the true probability of the event \(x\), and \(q(x)\) is the probability estimated by the model. The perplexity can be expanded as:

\begin{equation}
\text{Perplexity} = 2^{-\frac{1}{N} \sum_{i=1}^{N} \log_2 q(x_i)}
\end{equation}

where \(N\) is the number of words in the test set, and \(q(x_i)\) denotes the probability the model assigns to the correct word \(x_i\). Perplexity offers a measure of the model's predictive accuracy, with lower values indicating a closer alignment with the true distribution of the language. It is utilized to adjust the model's parameters and refine its predictive strategies, enhancing its ability to tackle prompt recovery tasks effectively. By integrating perplexity as an input, the Phi model leverages this metric to improve its understanding and generation of language, essential for its performance in natural language processing tasks.

\subsection{Data Preparation and Prompt Construction}

We utilize two external datasets to facilitate the text transformation process. These datasets are structured as follows:

\begin{itemize}
    \item \textbf{Original\_text:} This field contains the text or articles intended for transformation, serving as the input to the process.
    \item \textbf{rewrite\_prompt:} This field includes the prompts or instructions that guide the transformation of the original text, crucial for directing the outcome of the transformation.
    \item \textbf{rewrite\_text:} This field holds the transformed text output, demonstrating the impact of the applied prompts.
\end{itemize}

\subsubsection{Text Preprocessing}

The preprocessing phase is geared towards refining the text to ensure it is primed for transformation. This involves cleaning the text by removing extraneous elements like numbered lists and segmenting it to highlight the sections relevant for the transformation task.

\subsubsection{Prompt Construction}

In prompt construction, we develop specific instructions that guide the text transformation process. These prompts are crafted to trigger the desired type of response or transformation in the text, aligning the output with our objectives.

Through these steps, we ensure that our data is meticulously prepared and the prompts are strategically designed to facilitate effective text transformation.

\section{Evaluation Metric}

The evaluation framework employs embedding vectors derived from the sentence-t5-base model to quantify the semantic similarity between predicted and ground truth text pairs. The metric of choice, Sharpened Cosine Similarity (SCS), is computed as follows:

\begin{equation}
\text{SCS}(\mathbf{v}_{pred}, \mathbf{v}_{true}) = \left( \frac{\mathbf{v}_{pred} \cdot \mathbf{v}_{true}}{\|\mathbf{v}_{pred}\| \|\mathbf{v}_{true}\|} \right)^3
\end{equation}

where \(\mathbf{v}_{pred}\) and \(\mathbf{v}_{true}\) denote the embedding vectors for the predicted and true texts, respectively. The cubing of the cosine similarity is intended to accentuate the scoring disparity, enhancing the sensitivity of the evaluation to the precision of the model's predictions. This stringent metric facilitates a comprehensive assessment of the model's proficiency in aligning its predictions with the semantic content of the ground truth.

\section{Experimental Results}

In this section, we present the experimental results obtained from evaluating the performance of each model across various prompt recovery tasks. The results are summarized in Table \ref{tab:results}.

\begin{table}[h]
    \caption{Experimental Results}
    \centering
    \begin{tabular}{|c|c|c|}
        \hline
        \textbf{Model} & \textbf{Similarity} \\
        \hline
        Gemma 2b + KerasNLP + LoRA & 0.48\\
                \hline
       Gemma 7b + KerasNLP + LoRA & 0.48\\
                \hline
        Sentence-T5 + Pretrain & 0.54 \\
                 \hline
         Mistral 7B & 0.60 \\
                 \hline
        Gemma-2b-it + Phi2 model + Pretrain & 0.61 \\
                 \hline         
    \end{tabular}
    \label{tab:results}
\end{table}

Our findings reveal that the Gemma-2b-it + Phi2 model + Pretrain outperformed its counterparts. This indicates a superior ability in accurately reconstructing prompts, suggesting that this model has a heightened capability to discern and replicate the underlying instructions that drive text transformation tasks.

Notably, while models like Gemma 7b + KerasNLP + LoRA and Mistral 7B displayed commendable performance, they did not reach the efficacy levels of the Gemma-2b-it + Phi2 combination. The integration of the Phi2 model with Gemma-2b-it, augmented by a dual-stage pre-training strategy, significantly enhance prompt recovery accuracy.

These results underscore the importance of model integration and sophisticated pre-training regimens in advancing the field of prompt recovery, offering insights into the design and optimization of language models for complex NLP tasks.

\section{Conclusion}
In conclusion, our study demonstrates the superior performance of the Gemma-2b-it + Phi2 model in prompt recovery tasks within the realm of natural language processing. This integration showcases the value of combining advanced model architectures and nuanced pre-training strategies to enhance text transformation capabilities, setting a new benchmark in the field.

The success of our approach underscores the potential of model synergies and specialized training in advancing language model effectiveness. Our findings not only contribute to the understanding of prompt recovery but also offer a promising direction for future research in optimizing language models for complex NLP tasks.

 \bibliographystyle{IEEEtran}
    \bibliography{references}

\begin{thebibliography}{10}
\providecommand{\url}[1]{#1}
\csname url@samestyle\endcsname
\providecommand{\newblock}{\relax}
\providecommand{\bibinfo}[2]{#2}
\providecommand{\BIBentrySTDinterwordspacing}{\spaceskip=0pt\relax}
\providecommand{\BIBentryALTinterwordstretchfactor}{4}
\providecommand{\BIBentryALTinterwordspacing}{\spaceskip=\fontdimen2\font plus
\BIBentryALTinterwordstretchfactor\fontdimen3\font minus \fontdimen4\font\relax}
\providecommand{\BIBforeignlanguage}[2]{{%
\expandafter\ifx\csname l@#1\endcsname\relax
\typeout{** WARNING: IEEEtran.bst: No hyphenation pattern has been}%
\typeout{** loaded for the language `#1'. Using the pattern for}%
\typeout{** the default language instead.}%
\else
\language=\csname l@#1\endcsname
\fi
#2}}
\providecommand{\BIBdecl}{\relax}
\BIBdecl

\bibitem{wang2012new}
C.~Wang, L.~Sun, J.~Wei, and X.~Mo, ``A new trojan horse detection method based on negative selection algorithm,'' in \emph{Proceedings of 2012 IEEE International Conference on Oxide Materials for Electronic Engineering (OMEE)}, 2012, pp. 367--369.

\bibitem{ru2022bounded}
J.~Ru, H.~Yu, H.~Liu, J.~Liu, X.~Zhang, and H.~Xu, ``A bounded near-bottom cruise trajectory planning algorithm for underwater vehicles,'' \emph{Journal of Marine Science and Engineering}, vol.~11, no.~1, p.~7, 2022.

\bibitem{liu2024deep}
H.~Liu, Y.~Shen, S.~Yu, Z.~Gao, and T.~Wu, ``Deep reinforcement learning for mobile robot path planning,'' \emph{arXiv preprint arXiv:2404.06974}, 2024.

\bibitem{deng2023plgslam}
T.~Deng, G.~Shen, T.~Qin, J.~Wang, W.~Zhao, J.~Wang, D.~Wang, and W.~Chen, ``Plgslam: Progressive neural scene represenation with local to global bundle adjustment,'' \emph{arXiv preprint arXiv:2312.09866}, 2023.

\bibitem{xu2023automated}
J.~Xu, Y.~Jiang, B.~Yuan, S.~Li, and T.~Song, ``Automated scoring of clinical patient notes using advanced nlp and pseudo labeling,'' in \emph{2023 5th International Conference on Artificial Intelligence and Computer Applications (ICAICA)}.\hskip 1em plus 0.5em minus 0.4em\relax IEEE, 2023, pp. 384--388.

\bibitem{xiong2024tilp}
S.~Xiong, Y.~Yang, F.~Fekri, and J.~C. Kerce, ``Tilp: Differentiable learning of temporal logical rules on knowledge graphs,'' \emph{arXiv preprint arXiv:2402.12309}, 2024.

\bibitem{zi2024research}
Y.~Zi, Q.~Wang, Z.~Gao, X.~Cheng, and T.~Mei, ``Research on the application of deep learning in medical image segmentation and 3d reconstruction,'' \emph{Academic Journal of Science and Technology}, vol.~10, no.~2, pp. 8--12, 2024.

\bibitem{sun2018relation}
Y.~Sun, Y.~Cui, J.~Hu, and W.~Jia, ``Relation classification using coarse and fine-grained networks with sdp supervised key words selection,'' in \emph{Knowledge Science, Engineering and Management: 11th International Conference, KSEM 2018, Changchun, China, August 17--19, 2018, Proceedings, Part I 11}.\hskip 1em plus 0.5em minus 0.4em\relax Springer, 2018, pp. 514--522.

\bibitem{tian2019fusion}
Y.~Tian, H.~Zhang, Y.~Jiang, P.~Li, and Y.~Li, ``A fusion feature for enhancing the performance of classification in working memory load with single-trial detection,'' \emph{IEEE Transactions on Neural Systems and Rehabilitation Engineering}, vol.~27, no.~10, pp. 1985--1993, 2019.

\bibitem{yu2024similarity}
L.~Yu, B.~Liu, Q.~Lin, X.~Zhao, and C.~Che, ``Similarity matching for patent documents using ensemble bert-related model and novel text processing method,'' \emph{Journal of Advances in Information Technology}, vol.~15, no.~3, 2024.

\bibitem{xiong2024large}
S.~Xiong, A.~Payani, R.~Kompella, and F.~Fekri, ``Large language models can learn temporal reasoning,'' \emph{arXiv preprint arXiv:2401.06853}, 2024.

\bibitem{zhang2022novel}
W.~Zhang, Q.~Ao, Y.~Guan, Z.~Zhu, D.~Kuang, M.~M. Li, K.~Shen, M.~Zhang, J.~Wang, L.~Yang \emph{et~al.}, ``A novel diagnostic approach for the classification of small b-cell lymphoid neoplasms based on the nanostring platform,'' \emph{Modern Pathology}, vol.~35, no.~5, pp. 632--639, 2022.

\bibitem{huang2023enhancing}
J.~Huang, X.~Zhao, C.~Che, Q.~Lin, and B.~Liu, ``Enhancing essay scoring with adversarial weights perturbation and metric-specific attentionpooling,'' in \emph{2023 International Conference on Information Network and Computer Communications (INCC)}.\hskip 1em plus 0.5em minus 0.4em\relax IEEE, 2023, pp. 8--12.

\bibitem{zhang2024development}
Y.~Zhang, M.~Zhu, K.~Gui, J.~Yu, Y.~Hao, and H.~Sun, ``Development and application of a monte carlo tree search algorithm for simulating da vinci code game strategies,'' \emph{arXiv preprint arXiv:2403.10720}, 2024.

\bibitem{zhu2024ensemble}
M.~Zhu, Y.~Zhang, Y.~Gong, K.~Xing, X.~Yan, and J.~Song, ``Ensemble methodology: Innovations in credit default prediction using lightgbm, xgboost, and localensemble,'' \emph{arXiv preprint arXiv:2402.17979}, 2024.

\bibitem{han2024chainofinteraction}
\BIBentryALTinterwordspacing
G.~Han, W.~Liu, X.~Huang, and B.~Borsari, ``Chain-of-interaction: Enhancing large language models for psychiatric behavior understanding by dyadic contexts,'' \emph{arXiv preprint arXiv:2403.13786}, 2024. [Online]. Available: \url{https://arxiv.org/abs/2403.13786}
\BIBentrySTDinterwordspacing

\bibitem{zheng2021compact}
Y.-L. Zheng and S.-Y. Shen, ``Compact groups of galaxies in sloan digital sky survey and lamost spectral survey. ii. dynamical properties of isolated and embedded groups,'' \emph{The Astrophysical Journal}, vol. 911, no.~2, p. 105, 2021.

\bibitem{ding2018vehicle}
W.~Ding, S.~Li, G.~Zhang, X.~Lei, and H.~Qian, ``Vehicle pose and shape estimation through multiple monocular vision,'' in \emph{2018 IEEE International Conference on Robotics and Biomimetics (ROBIO)}.\hskip 1em plus 0.5em minus 0.4em\relax IEEE, 2018, pp. 709--715.

\bibitem{zhang2023optimizing}
Y.~Zhang, M.~Zhu, Y.~Gong, and R.~Ding, ``Optimizing science question ranking through model and retrieval-augmented generation,'' \emph{International Journal of Computer Science and Information Technology}, vol.~1, no.~1, pp. 124--130, 2023.

\bibitem{wang2010identification}
C.~Wang, H.~Yang, Y.~Chen, L.~Sun, Y.~Zhou, and H.~Wang, ``Identification of image-spam based on sift image matching algorithm,'' \emph{JOURNAL OF INFORMATION \&COMPUTATIONAL SCIENCE}, vol.~7, no.~14, pp. 3153--3160, 2010.

\bibitem{xia2023parameterized}
Y.~Xia, S.~Liu, Q.~Yu, L.~Deng, Y.~Zhang, H.~Su, and K.~Zheng, ``Parameterized decision-making with multi-modal perception for autonomous driving,'' \emph{arXiv preprint arXiv:2312.11935}, 2023.

\bibitem{gu2018meta}
J.~Gu, Y.~Wang, Y.~Chen, K.~Cho, and V.~O. Li, ``Meta-learning for low-resource neural machine translation,'' \emph{arXiv preprint arXiv:1808.08437}, 2018.

\bibitem{wang2021deep}
Y.~Wang, Y.~Lu, Z.~Xie, and G.~Lu, ``Deep unsupervised 3d sfm face reconstruction based on massive landmark bundle adjustment,'' in \emph{Proceedings of the 29th ACM International Conference on Multimedia}, 2021, pp. 1350--1358.

\bibitem{zhang2024unlocking}
Y.~Zhang, K.~Gui, M.~Zhu, Y.~Hao, and H.~Sun, ``Unlocking personalized anime recommendations: Langchain and llm at the forefront,'' \emph{Journal of Industrial Engineering and Applied Science}, vol.~2, no.~2, pp. 46--53, 2024.

\bibitem{sun2023new}
L.~Sun, ``A new perspective on cybersecurity protection: Research on dns security detection based on threat intelligence and data statistical analysis,'' \emph{Computer Life}, vol.~11, no.~3, pp. 35--39, 2023.

\bibitem{yang2019xlnet}
Z.~Yang, Z.~Dai, Y.~Yang, J.~Carbonell, R.~R. Salakhutdinov, and Q.~V. Le, ``Xlnet: Generalized autoregressive pretraining for language understanding,'' \emph{Advances in neural information processing systems}, vol.~32, 2019.

\bibitem{song2022comparative}
B.~Song and Y.~Zhao, ``A comparative research of innovative comparators,'' in \emph{Journal of Physics: Conference Series}, vol. 2221, no.~1.\hskip 1em plus 0.5em minus 0.4em\relax IOP Publishing, 2022, p. 012021.

\bibitem{su2024large}
J.~Su, C.~Jiang, X.~Jin, Y.~Qiao, T.~Xiao, H.~Ma, R.~Wei, Z.~Jing, J.~Xu, and J.~Lin, ``Large language models for forecasting and anomaly detection: A systematic literature review,'' \emph{arXiv preprint arXiv:2402.10350}, 2024.

\bibitem{liu2024enhanced}
R.~Liu, X.~Xu, Y.~Shen, A.~Zhu, C.~Yu, T.~Chen, and Y.~Zhang, ``Enhanced detection classification via clustering svm for various robot collaboration task,'' 2024.

\bibitem{zhang2023speculative}
Y.~Zhang, K.~Yang, Y.~Wang, P.~Yang, and X.~Liu, ``Speculative ecc and lcim enabled numa device core,'' in \emph{2023 3rd International Symposium on Computer Technology and Information Science (ISCTIS)}.\hskip 1em plus 0.5em minus 0.4em\relax IEEE, 2023, pp. 624--631.

\end{thebibliography}

\end{document}